\documentclass[11pt]{arxform}
\usepackage{graphicx}
\usepackage{multirow}
\usepackage{todonotes}
\usepackage{stmaryrd}
\usepackage{amsmath}
\usepackage{amssymb}
\usepackage{hyperref}
\usepackage{subfig}

\usepackage{algorithm}
\usepackage{algorithmic}

\usepackage{xcolor}
\usepackage{hyperref}

\usepackage{times}
\usepackage{multirow,color}
\usepackage{amsmath} 
\renewcommand{\vec}[1]{\boldsymbol{#1}}

\usepackage{algorithm}
\usepackage{algorithmic}
\usepackage{stmaryrd}
\usepackage{amsfonts}
\usepackage{dsfont}
\usepackage{tabularx}

\begin{document}
	\title{Learning Tversky Similarity\thanks{Draft of an article in Proc.\ IPMU 2020, International Conference on Information Processing and Management of Uncertainty in Knowledge-Based Systems.}}
	%
	%
	\author{Javad Rahnama and Eyke H{\"u}llermeier}
	%
	%
	
	\institute{Paderborn University\\
		Heinz Nixdorf Institute and Department of Computer Science\\
		Intelligent Systems and Machine Learning Group\\
		\email{javad.rahnama@uni-paderborn.de, eyke@upb.de}}

\maketitle

	\begin{abstract}
	In this paper, we advocate Tversky's ratio model as an appropriate basis for computational approaches to semantic similarity, that is, the comparison of objects such as images in a semantically meaningful way. We consider the problem of learning Tversky similarity measures from suitable training data indicating whether two objects tend to be similar or dissimilar. Experimentally, we evaluate our approach to similarity learning on two image datasets, showing that is performs very well compared to existing methods.\\
	\textbf{Key words}: Similarity, machine learning, semantic features, image data
	\end{abstract}

\section{Introduction}

Similarity is an important cognitive concept and a key notion in various branches of artificial intelligence, including case-based reasoning \cite{Richter95}, information retrieval \cite{Chechik_2010}, machine learning \cite{Chen_2009}, data analysis \cite{Strehl_2000} and data mining \cite{bouc_si08}, among others. Specified as a numerical (real-valued) function on pairs of objects, it can be applied in a rather generic way for various problems and purposes \cite{leso_sm09}. In particular, similarity is of great interest for structured objects such as text and images. In such domains, similarity measures are normally not defined by hand but learned from data, i.e., they are automatically extracted from sample data in the form of objects along with information about their similarity.

In the image domain, numerous methods of that kind have been proposed, based on different types of feature information, including visual \cite{Deselaers_2011,Wang_2014} and semantic features \cite{Pedersen_2004}, and exploiting different types of measurements, such as non-metric \cite{Garcia_2019,Chechik_2010} and metric ones \cite{Kulis_2013}. Most common is a geometrical model of similarity that relies on an embedding of objects as points in a suitable (often high-dimensional) vector space, in which similarity between objects is inversely related to their distance. As popular examples of (dis-)similarity measures in this field, let us mention the Euclidean and the Cosine distance \cite{qian2004similarity}.

Somewhat surprisingly, another concept of similarity, the one put forward by Tversky \cite{tversky_1977}, appears to be less recognized in this field, in spite of its popularity in psychology and cognitive science. 
Tversky argued that the geometrical model of similarity is not fully compatible with human perception, and empirically demonstrated that human similarity assessment does not obey all properties implied by this model, such as minimality, symmetry, and triangle inequality \cite{tversky_1982}. Due to its cognitive plausibility, we hypothesize that Tversky similarity could provide a suitable basis for mimicking human similarity assessment in domains such as \emph{art images} \cite{lang2018attesting}. For example, an art historian will probably find the copy of a painting more similar to the original than the original to the copy, thereby violating symmetry. Not less importantly, Tversky similarity is a potentially interpretable measure, which does not merely produce a number, but is also able to explain why two objects are deemed more or less similar. In other words, it is able to fill the \textit{semantic gap} between the extracted visual information and a human's interpretation of an image. This feature of \emph{explainability} is of critical importance in many applications \cite{Smeulders_2000}. Last but not least, the similarity also exhibits interesting theoretical properties; for example, see \cite{cole_as19} for a recent analysis from the perspective of measurement theory.

In this paper, we elaborate on the potential of Tversky similarity as an alternative to existing measures, such as Euclidean and Cosine distance, with a specific focus on the image domain. In particular, we consider the problem of learning Tversky similarity from data, i.e., tuning its parameters on the basis of suitable training data, a problem that has not received much attention so far. As a notable exception we mention \cite{baio_wa12}, where the problem of learning the importance of features (and feature combinations) for a generalized Jaccard measure (a special case of the  Tversky measures) on the basis of training data is addressed, so as to achieve optimal performance in similarity-based (nearest neighbor) classification; to this end, the authors make use of stochastic optimization techniques (particle swarm optimization and differential
evolution).
On the other side, it is also worth noticing that the problem of learning Tversky similarity is in general different from the use of the Tversky measure as a loss function in machine learning \cite{sale_tl17,abra_an19}. Here, the measure serves as a means to accomplish a certain goal, namely to learn a classification model (e.g., a neural network) that achieves a good compromise between precision and recall, whereas is our case, the measure corresponds to the sought model itself.

The paper is organized as follows. After a brief review of related work in the next section, we recall Tversky's notion of similarity based on the ratio model in Section~\ref{tversky_sim}. In Section~\ref{learning}, we then propose a method for learning this similarity from suitable training data. Finally, we present experimental results in Section~\ref{experiments}, prior to concluding the paper in Section~\ref{conclusion}.

\section{Related Work} \label{related_work}

\subsection{Image Similarity}

Image similarity and its quantitative assessment in terms of similarity measures strongly depends on the image representation. Numerous approaches have been presented to extract different types of representations based on visual and semantic features of images ~\cite{Deselaers_2011,Liu_2007}. Most of the state-of-the-art methods for extracting visual features are based on deep neural networks, which produce a variety of features, ranging from low level features in the early layers to more abstract features in the last layers \cite{krizhevsky_2012,Yue_2015}. In general, however, similarity is a complex concept that can not be derived from visual features alone. Therefore, some studies have exploited the use of prior knowledge as well as intermediate or high-level representation of features to capture similarity in specific applications \cite{Datta_2008,Deselaers_2011,Balt_2018}.

The similarity or distance on individual features is commonly combined into an overall assessment using measures such as weighted Euclidean or weighted Cosine distance, whereas Tversky similarity is much less used in the image domain in this regard. There are, however, a few notable exceptions. For example, Tversky similarity is used in the context of image retrieval in \cite{santini} and for measuring the similarity between satellite images in \cite{tang}.

\subsection{Metric Learning}

As already said, the notion of similarity is closely connected to the notion of distance. Whether relationships between objects are expressed in terms of similarity or distance is often a matter of taste, although small differences in the respective mathematical formalizations also exist. In the literature, distance seems to be even a bit more common than similarity. In particular, there is large body of literature on distance (metric) learning \cite{Kulis_2013}. Most distance learning methods focus on tuning the weights of attributes in the Mahalanobis distance or weighted Euclidean distance. As training information, these methods typically use constraints on pairwise or relative relations among objects \cite{Hermans_2017,lecun_2005}.

 \section{Tversky Similarity}
 \label{tversky_sim}

Tversky suggested that humans perceive the similarity between objects based on contrasting (the measure of) those features they have in common and those on which they differ \cite{tversky_1977}. Moreover, he suggested that more attention is paid to the shared than to the distinctive features. Thus, in his  (feature-matching) model, an object is represented as a set of meaningful features, and similarity is defined based on suitable set operations. 

\subsection{Formal Definition}

More formally, consider a set of objects $\mathcal{X}$ and a finite set of features $\mathcal{F}= \{ f_1, \ldots , f_m \}$. Each feature is considered as a binary predicate, i.e., a mapping $f_i : \, \mathcal{X} \longrightarrow \{ 0, 1 \}$, where $f_i(\vec{x}) = 1$ is interpreted as the presence of the $i^{th}$ feature for the object $\vec{x}$, and  $f_i(\vec{x}) = 0$ as the absence. Thus, each object $\vec{x} \in \mathcal{X}$ can be represented by the subset of features it exhibits:
$$
F(\vec{x}) = \big\{ f_i \, \vert \, f_i(\vec{x}) = 1 , \, i \in \{1, \ldots , m \} \big\} \subseteq \mathcal{F}
$$
Tversky similarity, in the form of the so-called ratio model, is then defined as a function $S_{\alpha,\beta}: \, \mathcal{X}^2 \longrightarrow [0,1]$ as follows: 
\begin{equation}
\label{eq_tv}
S_{\alpha,\beta}(\vec{x},\vec{y}) = \frac{g( \, |A \cap B| \, )}{g( \, |A \cap B| \, ) + \alpha\, g( \, |A  \setminus B| \, ) + \beta \, g( \, |B  \setminus A| \, )} \, ,
\end{equation}
where $A= F(\vec{x})$, $B= F(\vec{y})$, and $g$ is a non-negative, increasing function $\mathbb{N}_0 \longrightarrow \mathbb{N}_0$; in the simplest case, $g$ is the identity measuring set cardinality. According to (\ref{eq_tv}), Tversky similarity puts the number of features that are shared by two objects in relation to the number of relevant features, where relevance means that a feature is present in at least one of the two objects\,---\,features that are absent from both objects are completely ignored, which is an important property of Tversky similarity, and distinguishes it from most other similarity measures.

The coefficients $0 \leq \alpha, \beta \leq  1$ in (\ref{eq_tv}) control the impact of the distinctive features as well as the asymmetry of the measure. The larger these coefficients, the more important the distinctive feature are. For $\alpha = \beta$, the similarity measure is symmetric, i.e., $S_{\alpha,\beta}(\vec{x},\vec{y}) = S_{\alpha,\beta}(\vec{y},\vec{x})$ for all $\vec{x}, \vec{y} \in \mathcal{X}$, for $\alpha \neq \beta$ it is asymmetric. Important special cases include the Jaccard coefficient ($\alpha = \beta = 1$) and the Dice similarity ($\alpha = \beta = 1/2$).

\subsection{Feature Weighting}

The Tversky similarity (TS) measure (\ref{eq_tv}) implicitly assumes that all features have the same importance, which might not always be true. In fact, $g$ is only a function of the cardinality of feature subsets, but ignores the concrete elements of these subsets. In other words, only the number of shared and distinctive features is important, no matter what these features are. 

A natural generalization of (\ref{eq_tv}) is  a weighted variant of Tversky similarity (WTS), in which each feature $f_i$ is weighted by some $w_i \in [0,1]$:
\begin{equation}
\label{weight_tv}
   S_{\alpha,\beta,\vec{w}}(\vec{x},\vec{y}) = 
   \frac{\sum_{i=1}^m w_i f_i(\vec{x})f_i(\vec{y})}{\sum_{i=1}^m w_i \Big( 
   f_i(\vec{x})f_i(\vec{y}) + \alpha  f_i(\vec{x}) \bar{f}_i(\vec{y}) + \beta  \bar{f}_i(\vec{x})f_i(\vec{y})
   \Big) } \, ,
\end{equation}
with $\bar{f}_i(\cdot) = 1 - f_i(\cdot)$.
Thus, in addition to $\alpha$ and $\beta$, this version of Tversky similarity is now parametrized by a weight vector $\vec{w} = (w_1, \ldots , w_m)$.

\subsection{Semantic Similarity}
\label{sec:semantic}

As already said, we believe that Tversky similarity may offer a semantically meaningful and cognitively plausible tool for object comparison, especially in the image domain\,---\,provided the $f_i$ are ``semantic features'', that is, features with a meaningful semantic interpretation. In the image domain, such features are normally not given right away. Instead, meaningful (high-level) features have to be extracted from ``low-level'' feature information on the pixel level.

\begin{figure}
\begin{center}
\includegraphics[scale=0.35]{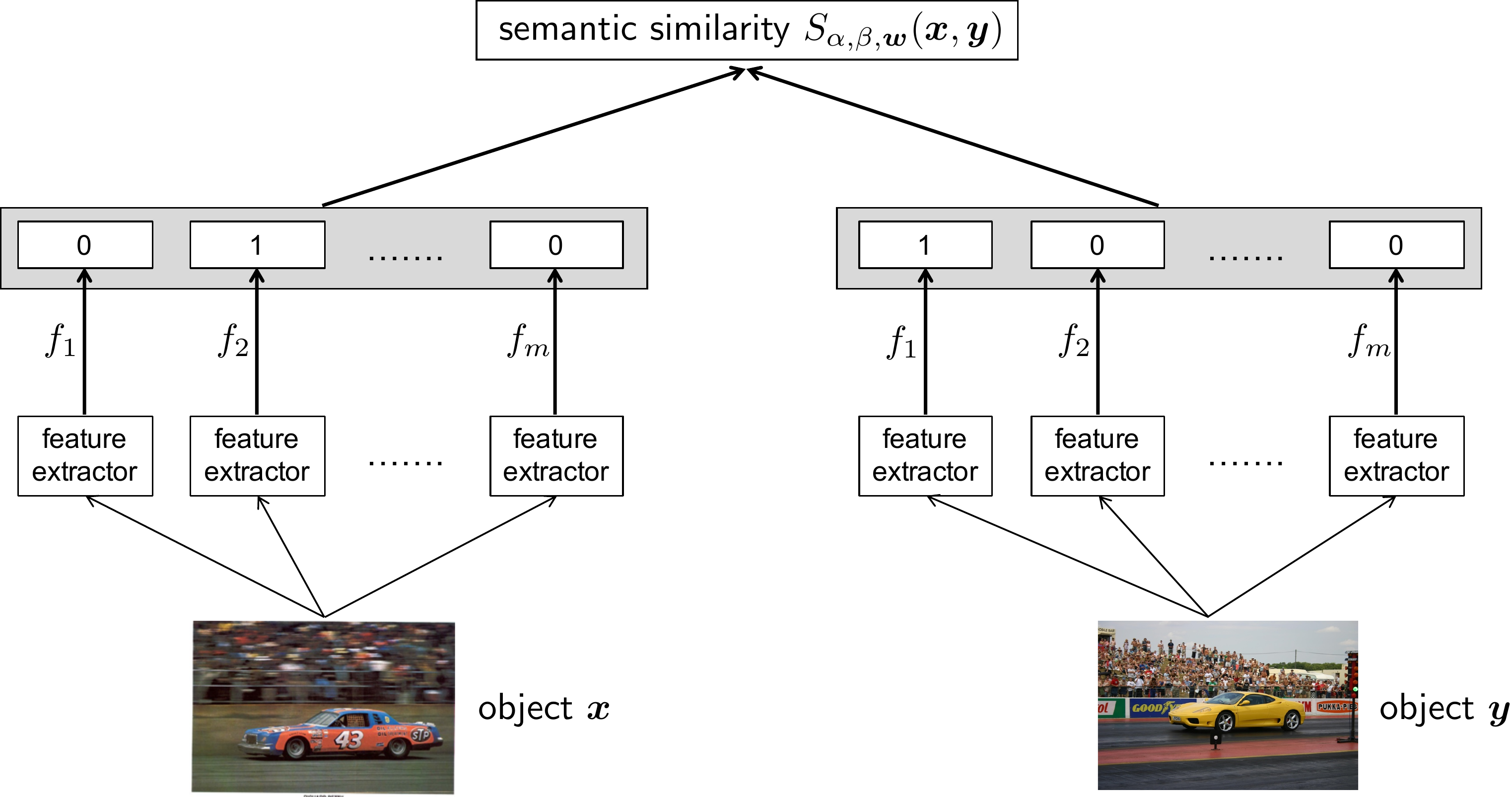}
\caption{Two-level architecture for semantic similarity.}
\label{fig:sesi}
\end{center}
\end{figure}

Thus, what we envision is an approach to image similarity based on a two-level architecture, in which semantic features are extracted from images in the first step, and these features are then used to specify similarity in the second step (cf.\ Fig.\ \ref{fig:sesi}). While the features could be predefined by a human domain expert, the feature extraction itself should be accomplished automatically based on suitable image processing techniques. 

For example, a human expert may know that, in a certain context, the presence of water on an image is a relevant feature. Water might then be included as a feature $f_i$ in $\mathcal{F}$. The function $f_i$ itself is a mapping that assumes an image (represented as a set of pixels) as an input and returns $0$ or $1$ as an output, depending on whether water is visible on the image or not. It could be realized, for example, by a neural network that has been trained to recognize water on images. Obviously, feature descriptors obtained in this way are never perfect, so that feature descriptions $F(\vec{x})$ will be ``noisy'' in practice.

\section{Learning Tversky Similarity}
\label{learning}
In this section, we address the problem of \emph{learning} the Tversky similarity, that is, tuning the parameters of the measure so as the optimally adapt it to a concrete application at hand. To this end, we assume suitable training data to be given, which informs about the similarity or dissimilarity between objects. More specifically, we assume training data of the following form:
\begin{equation}\label{eq:data}
\mathcal{D} = \big\{ (F(\vec{x}_n), F(\vec{y}_n ), s_n ) \big\}_{n=1}^N \subset \mathcal{X} \times \mathcal{X} \times \{ 0,1 \}  
\end{equation}
Each training example is a triplet $(F(\vec{x}_n) , F(\vec{y}_n) , s_n )$, where $s_n \in \{0,1 \}$ indicates whether the two objects $\vec{x}_n$ and $\vec{y}_n$ are considered similar or not. Thus, instead of precise numerical similarity degrees, the feedback is binary and only provides a rough indication. Note that the objects in the training data are already represented in terms of a feature description, i.e., we disregard the step of feature extraction here and completely focus on specifying the Tversky similarity.

\subsection{Contrastive Loss}

A common approach to learning similarity or distance functions is to minimize a suitable loss function on the training data $\mathcal{D}$. Here, we make use of the contrastive loss  \cite{lecun_2005}, which has been employed quite successfully in the image domain. This loss function compares a ``predicted'' similarity $\hat{s} = S_{\alpha,\beta,\vec{w}}( \vec{x}, \vec{y}) \in [0,1]$ with given feedback $s \in \{0,1 \}$ as follows: 
\begin{equation}\label{eq:cl}
L_m(s , \hat{s}) = s (1- \hat{s}) + (1-s) \max \big(m - 1 + \hat{s}  , 0 \big) \, ,
\end{equation}
where $m$ is a margin (and a parameter of the loss). Thus, if $s=1$ (objects $\vec{x}$ and $\vec{y}$ are considered similar), the loss is given by $1- \hat{s}$ and increases linearly with the distance from the ideal prediction $\hat{s} = 1$. If $s=0$, the loss is specified analogously, except that a loss of 0 only requires $\hat{s} \leq 1-m$ (instead of $\hat{s}=0$). Thus, positive and negative examples are treated in a slightly different way: A high (predicted) similarity on negative pairs is penalized less strongly than a low similarity on positive pairs. 
 This could be meaningful, for example, when the feedback $s$ is obtained from class membership in a classification or clustering context (as will be the case in our experiments) with many classes, some of which may (necessarily) overlap to a certain extent.

\subsection{Loss Minimization}

Training the weighted Tversky similarity (WTS) essentially consists of minimizing the total loss on the training data, i.e., finding
\begin{equation}
\label{eq:loss}
(\alpha^*, \beta^*, \vec{w}^*) = \operatorname*{argmin}_{ \alpha, \beta, \vec{w}} \sum_{n=1}^N L_m(s_n, S_{ \alpha, \beta, \vec{w}}(\vec{x}_n, \vec{y}_n)) 
\end{equation}
Besides, we consider two restricted variants of this problem: Learning the (unweighted) TS measure comes down to omitting the weight vector $\vec{w}$ from (\ref{eq:loss}), and enforcing symmetry to optimizing a single parameter $\alpha$ instead of two parameters $\alpha$ and $\beta$. 

Our learning algorithm (see Algorithm \ref{alg:learning}) is quite simple and essentially based on gradient descent optimization. Gradients for the parameter updates are computed on mini-batches. To avoid overfitting of the training data, we take a part of the data for validation and apply an early stopping technique. More specifically, we stop the training process as soon as the accuracy on the validation data decreases by more than $0.01$ in $20$ consecutive iterations.

\begin{algorithm}
\definecolor{dark-gray}{rgb}{0.25,0.25,0.25}

    \caption{Learning Algorithm}
    \label{alg:learning}
    \begin{algorithmic}[1]
       \REQUIRE maximum iteration $MT$, batch size $ B $, margin $ m $,  training data $\mathcal{D}$
       \ENSURE $ \alpha , \beta, \vec{w}$
        
        \STATE split $\mathcal{D}$ into training data $\mathcal{D}_{train}$ and validation data $\mathcal{D}_{val}$
        \STATE randomly initialize parameters $\alpha,\beta, \vec{w}$
        
        \WHILE{ ($t<MT$) and (stopping criterion $=$ false)}
        \STATE sample mini-batch uniformly at random from similar and dissimilar pairs
              \STATE compute $S_{\alpha, \beta, \vec{w}}$ according to (\ref{weight_tv})
    
            \STATE update parameters by minimizing (\ref{eq:loss})
      
              \STATE test stopping condition on $\mathcal{D}_{val}$
        \ENDWHILE
        
    \end{algorithmic}
\end{algorithm}

\section{Experiments}\label{experiments}

In this section, we present the results of our experimental study. Before doing so, we first describe the data we used for the experiments, the experimental setting, and the baseline methods we compared with.

\subsection{Data}

Since the extraction of semantic features (cf. Section \ref{sec:semantic}) is beyond the scope of this paper, we collected two image datasets for which this information, or at least information that can be interpreted as such, is already given.

The \textbf{a-Pascal-VOC2008} data \cite{Farhadi_2009} consists a total of 4340 images, which are split into 2113 training and 2227 test images. Each image is labeled with one of 32 class categories, such as ``horse''. Moreover, each image is described by 64 additional binary attributes, which we interpret as semantic features. Mostly, these features characterize the presence of certain objects on the image, such as saddle, tail, or snout.

The \textbf{Sun attributes} dataset \cite{patterson_2014} includes 14340 images equally distributed over 717 categories, such as ``forest road'' or ``mountain path''. Each image is also described by 102 attributes, such as ``snow'', ``clouds'', ``flowers'', etc., which describe images in a semantically meaningful way. 
These features are attached by 3 human annotators, and we assume the presence of each attribute in an image if it is attached by at least one of them.

\subsection{Experimental Setting}

We split the Sun attribute dataset into training data, validation data, and test data with a ratio of 70/10/20. The validation data is used to fine-tune parameters like the margin $m$ of the contrastive loss. 
After each iteration of the training procedure (cf.\ Algorithm \ref{alg:learning}), the model is evaluated on validation data, and the model with highest performance is stored. We consider the same setting for the a-Pascal-VOC dataset, except for adopting the predefined train/test split. For this data, we extract $10\%$ of the training data for validation.

Training examples for similarity learning (i.e., triplets of the form $(\vec{x}, \vec{y}, s)$) are extracted from the set of images belonging to the training data as follows: A pair of images $\vec{x}, \vec{y}$ is considered as similar ($s=1$) if both images have the same class label, and otherwise as dissimilar ($s=0$). To facilitate learning and have balanced batches, similar and dissimilar pairs of images are sampled with the same probability of $0.5$ (and uniformly within the set of all similar pairs and all dissimilar pairs, respectively).

Once a similarity measure $S$ (or distance measure) has been learned, its performance is evaluated on the test data. To this end, we generate data in the form of triplets $(\vec{x}, \vec{y}, s)$ in the same way as we did for training. The similarity measure is then used as a threshold classifier $(\vec{x}, \vec{y}) \mapsto \llbracket S(\vec{x}, \vec{y}) > t \rrbracket$ with a threshold $t$ tuned on the validation data, and evaluated in terms of its classification rate (percentage of correct predictions) and F1 measure. Eventually, the average performance is determined over a large set of randomly selected triplets (4M in the case of the a-Pascal-VOC2008 data and 6M in the case of Sun attributes), so that the estimation error is essentially 0.

\subsection{Methods}

We train both variants of the Tversky similarity, the unweighted (TS) and the weighted one (WTS). Since the ``ground-truth'' similarity in our data is symmetric by construction, we restrict to the respective symmetric versions ($\alpha = \beta$). We train unweighted Tversky similarity and weighted Tversky similarity using stochastic gradient descent with Nestrov's momentum~\cite{sutskever2013} (learnin rate 0.01) and Adagrad~\cite{duchi2011adaptive} (learning rate 0.01), respectively.

As baselines for comparison, we learn two common similarity/distance metrics, namely the weighted Euclidean distance and weighted Cosine distance. Both distances are trained using Adam optimization~\cite{kingma2014adam} (with learning rate 0.001), in very much the same ways as the Tversky similarity. To prevent overfitting, the contrastive loss is combined with L1 and L2 regularization terms.

Moreover, we include LMNN \cite{Weinberger_2006} and a modified Siamese network (Siamese Net Semantic) based on \cite{Garcia_2019}. The modified Siamese network consists of two blocks: two feature extraction blocks with shared weights for a pair of objects and a non-metric similarity block. A feature extraction block maps the input semantic feature vector into an embedding space, in which similar objects have small and dissimilar objects a larger distance. The non-metric similarity block predicts the similarity score that indicates the degree of similarity or dissimilarity of the input pair.  
 
Since the number of input (semantic) features for the feature extraction block is relatively low in our experiments with the Sun attribute dataset, we only use three fully connected layers of size $[64, 32, 16]$ and activation functions [Relu, Relu, Sigmoid], respectively. Correspondingly, the dimensions of two fully connected layers are $[32, 16]$ with [Relu, Sigmoid] as activation functions for the a-Pascal-VOC2008 dataset. The non-metric network consists of an L1 distance part and two fully connected layers. In the L1 distance part, we calculate the L1 distance between features of the object pair, produced by the feature extraction blocks. The dimensions of two fully connected layers are $[16, 1]$ with [Relu, Sigmoid] as activation functions, respectively. The parameters of the modified Siamese network are learned by minimizing a combination of two common loss functions using back propagation and the Adam optimization method \cite{kingma2014adam} with learning rate 0.01. The contrastive loss is exploited to pull the representation of similar images closer and push the dissimilar ones apart. Moreover, since the evaluation is eventually done on a binary classification task, we combine the contrastive loss with the cross-entropy loss to improve the classification accuracy.

To show the effectiveness of our method in obtaining semantic similarity among images, we also train the modified Siamese network based on only visual features (Siamese Net Visual). The inputs of this network are the original images, and the output is a similarity prediction that indicates whether the two input images are similar or not. In the feature extraction block, we extract high-level features from the pre-trained Inception-V3 \cite{szegedy2016} followed by a flatten layer, batch normalization, and two fully connected layers with dimensions $[256, 128]$ and activation functions [Relu, Sigmoid]. We also use the same non-metric similarity block and optimization method as explained above with a learning rate 0.01.

\subsection{Results}

The results are summarized in Tables \ref{tab_pas} and \ref{tab_sun}. As can be seen, Tversky similarity performs very well and is highly competitive. Somewhat surprisingly, the performance of the simple unweighted version of Tversky similarity is already extremely strong. It can still be improved a bit by the weighted version, but not very much. 

The Euclidean (and likewise the Cosine) distance performs quite poorly. As a possible explanation, note that the Euclidean distance is not able to ignore presumably irrelevant features that occur in none of the two objects (images) at hand. Since the distance between the corresponding feature values is 0, such features contribute to the similarity of the objects (e.g., the simultaneous absence of, say, trees on two images contributes to their similarity). This is certainly a strength of the Tversky similarity (and part of its motivation).

\begin{table}
\caption{Performance on the a-Pascal-VOC2008 dataset.}\label{tab_pas}
\begin{center}
\begin{tabular}{lc@{\quad}c}
\hline
method & classification rate & F1 measure \\
\hline
weighted Euclidean &  0.65 & 0.74\\
weighted Cosine &  0.81 & 0.78\\
LMNN~\cite{Weinberger_2006} &  0.78 & 0.81\\
Siamese Net Semantic~\cite{Garcia_2019} &  0.82 & \textbf{0.84}\\
Tversky (TS) &  0.82 & 0.82\\
weighted Tversky (WTS) &  \textbf{0.83} & \textbf{0.84}\\
\hline
\end{tabular}
\end{center}
\end{table}

\begin{table}
\caption{Performance on the Sun attribute dataset.}\label{tab_sun}
\begin{center}
\begin{tabular}{lc@{\quad}c}
\hline
method  & classification rate & F1 measure \\
\hline
weighted Euclidean  & 0.76 & 0.77\\
weighted Cosine & 0.78 & 0.80\\
LMNN~\cite{Weinberger_2006} &  0.79 & 0.81\\
Siamese Net Semantic~\cite{Garcia_2019} &  0.79 & 0.81\\
Siamese Net Visual~\cite{Garcia_2019} & 0.74 & 0.78 \\
Tversky (TS) & 0.80 & 0.81\\
Weighted Tversky (WTS) & \textbf{0.81} & \textbf{0.83}\\

\hline
\end{tabular}
\end{center}
\end{table}

\section{Conclusion}\label{conclusion}

This paper presents first steps toward learning Tversky similarity, i.e., machine learning methods for tuning the parameters of Tversky's similarity model to training data collected in a concrete application context. To the best of our knowledge, such methods do not exist so far, in spite of the popularity of Tversky's model. The experimental results we obtained for image data so far, even if preliminary, are quite promising.  

There are various directions for extending the approach presented in this paper, which ought to be addressed in future work:
\begin{itemize}
\item Our learning algorithm implements a rather plain solution and essentially applies a general purpose learning technique (a gradient-based optimization) to a specific loss function. More sophisticated methods, specifically tailored to the Tversky similarity and exploiting properties thereof, promise to improve efficiency and perhaps even boost performance. In this regard, we plan to elaborate on different ideas, such as alternating optimization and correlation analysis to judge feature importance.

\item In practical applications, other types of training data may become relevant. An interesting example is relative similarity information of the form ``object $\vec{x}$ is more similar to $\vec{y}$ than to $\vec{z}$''. Devising methods for learning from data of that kind is another important topic of future work.

\item Likewise, other loss functions and performance metrics should be considered, both for training and evaluation. Specifically relevant are ranking measures from information retrieval, because similarity is often used for the purpose of object retrieval (e.g., ranking images stored in a database in decreasing order of their similarity to a query image).

\item
Further generalizations of the Tversky similarity itself could be considered as well, for example using fuzzy instead of binary features \cite{tolias_2001,cole_fw17,cole_fs19}.

\end{itemize}

\subsubsection*{Acknowledgements}
This work was supported by the German Research Foundation (DFG) under grant HU 1284/22-1. The authors strongly profited from discussions with the collaborators Ralph Ewerth, Hubertus Kohle, Stefanie Schneider, and Matthias Springstein.

%
%
%

\end{document}